\documentclass{article}

\usepackage[final, nonatbib]{neurips_2019}

\usepackage[utf8]{inputenc} 
\usepackage[T1]{fontenc}    
\usepackage{hyperref}       
\usepackage{url}            
\usepackage{booktabs}       
\usepackage{amsfonts}       
\usepackage{nicefrac}       
\usepackage{microtype}      
\usepackage{amsmath}
\usepackage{graphicx}
\usepackage{subcaption}
\usepackage{multirow}

\title{On the Interpretability and Evaluation of Graph Representation Learning}

\author{%
  Antonia Gogoglou \\
  Capital One\\
  McLean, VA 22102 \\
  \texttt{antonia.gogoglou@capitalone.com} \\
  \And 
  C. Bayan Bruss \\
  Capital One\\
  McLean, VA 22102 \\
  \texttt{bayan.bruss@capitalone.com} \\
  \And 
  Keegan E. Hines \\
  Capital One\\
  McLean, VA 22102 \\
  \texttt{keegan.hines@capitalone.com} \\
}

\begin{document}

\maketitle

\begin{abstract}
  With rising interest in graph representation learning, a variety of approaches have been proposed to effectively capture a graph’s properties. While these approaches have improved performance in graph machine learning tasks compared to traditional graph mining techniques, they are still perceived as {\it black-box} techniques with limited insights into the information encoded in these representations. In this work, we explore methods to interpret node embeddings and propose the creation of a robust evaluation framework for comparing graph representation learning algorithms and hyperparameters. We test our methods on graphs with different properties and investigate the relationship between embedding training parameters and the ability of the produced embedding to recover the structure of the original graph in a downstream task. 
\end{abstract}

\section{Introduction}
Graphs play a key role in many machine learning tasks providing the structured information needed to learn meaningful patterns and generate predictive models. However, it is challenging to represent complex structures like graphs in an expressive and efficient way that they can be fed into machine learning applications. Advances in the field of Graph Representation Learning \cite{hamilton2017representation, goyal2018graph} appear to provide a mapping that embeds nodes, or entire graphs, as dense low dimensional vectors.
Recently proposed approaches such as DeepWalk \cite{perozzi2014deepwalk}, LINE \cite{tang2015line}, node2vec \cite{grover2016node2vec}, GCNs \cite{kipf2016semi} and GraphSage \cite{hamilton2017inductive} treat this mapping as a machine learning task itself and aim to optimize it so that relationships in the embedding space accurately reflect the topology of the original graph.

A common categorization distinguishes between {\it shallow} and {\it deep} node embeddings. Shallow embeddings rely on first- or higher-order proximity derived from the original graph, often via random walks, to provide the context of a node and inform its representation.  Deep learning approaches include Graph Convolutional Networks (GCNs) and Message Passing Neural Networks (MPNNs) which extend the concept of convolution to describe a node as a function of its neighborhood. Regarding the objective to optimize during training of the embeddings, unsupervised approaches optimize for link reconstruction, supervised approaches for an externally assigned node label and semi-supervised operate on a subset of labeled nodes.

Given that different embedding approaches optimize for different objectives and operate on different input, it is expected that there is not a single "one-fits-all" node embedding technique. 
Recent work has focused on evaluating graph representation learning techniques with regards to their ability to distinguish graph properties \cite{dalmia2018towards, xu2018powerful}. In this direction, we investigate the interpretability of node embeddings and propose an evaluation framework that answers the following questions:
\begin{itemize}
    \item What information do node embeddings express and can we derive metrics to quantify their properties?
    \item How can we evaluate node embeddings with or without external labels and is there a single approach that maximizes performance across all tasks?
    \item Can complicated structures of the original graph be captured in embeddings trained on the local context around a node?
\end{itemize}
The rest of the paper is organized as follows: Section \ref{Methods} describes our proposed methodology, while Section \ref{Results} shows the results of our experiments and concludes the article.

\subsection{Methods}
\label{Methods}

\subsubsection{Interpretability}
In graph representation learning, nodes are typically embedded into a fixed D dimensional vector space (where D is a hyperparameter) Theoretically, the space is as condensed of a representation as we can get, without loss of information. This indicates that an {\it interpretable} embedding dimension would be highly associated with a particular feature of the original graph, a so-called disentangled representation \cite{higgins2017beta, bouchacourt2018multi, locatello2018challenging}. In NLP these features are often expressed in the form of semantic categories of words \cite{csenel2018semantic, park2017rotated}. In the case of graphs such categories can be derived from extrinsic or intrinsic sources, with the former being categories or labels assigned externally to nodes while the latter refers to groups found in the decomposition of the original graph (e.g. communities or partitions).

We define an {\it Interpretability Score} adapted from \cite{csenel2018semantic} for each dimension and each group of nodes:

\begin{equation}
    IS_{top(d,l)} = \frac{|C_l \cap top_k(E_d)|}{|C_l|} \times 100\  \\
    \ \ \ \  IS_{bottom(d,l)} = \frac{|C_l \cap bottom_k(E_d)|}{|C_l|} \times 100
\end{equation}

where $C_l$ is the $l_{th}$ group of nodes  and $E_d$ is the $d_{th}$ embedding dimension, while $k$ is a hyperparameter set equal to the cardinality of $C_l$ for our experiments. Interpretability scores are produced for both the top and bottom items of each embedding dimension and they can be aggregated by taking the maximum or average. Thereafter, scores are aggregated either across multiple groups to get the score for a single embedding dimension or across embedding dimensions to obtain per group scores.

\begin{equation}
    IS_d = \underset{d=0 \ to \ D}{agg_1}(agg_2(IS_{top(dl)}, IS_{bottom(d,l)}))
    \ \ \ \ \ 
    IS_l = \underset{l=0 \ to \ L}{agg_1}(agg_2(IS_{top(dl)}, IS_{bottom(d,l)}))
    \ \ \ \ \ 
\end{equation}
If the top nodes in the positive or negative direction of an embedding dimension are highly associated with a particular node category and at the same time have lower overlap with the rest of the categories, then the interpretability of this dimension is strong.

\subsubsection{Embedding approaches and Datasets}
In random-walk based embedding models, there are two general components, a system for generating long random walks (with some variants depending on the model), and a shallow one layer neural network skip-gram model. Each of the components contains a set of hyperparameters out of which the most commonly reported one is embedding dimensionality. 

To investigate the proposed evaluation methods we use three datasets: one coming from the financial sector ({\it Brand Level Merchants - BLM}) \cite{bruss2019deeptrax} and two from the social networks sphere ({\it BlogCatalog}) and ({\it Flickr}) \cite{tang2009relational}. The BlogCatalog dataset contains friendship connections between bloggers. Additionally, it contains labels for each node referring to 39 categories the bloggers could be affiliated with. Similarly, Flickr data contains links between users of the Flickr board and 195 categories users can be associated with. The Brand Level Merchant dataset is constructed from credit card transaction logs. By taking any two transactions that share an account within a specified time window, a set of merchant pairs, meaning walk lengths equal to 2, are generated.  For all datasets we generate embeddings using the GENSIM implementation of word2vec \cite{rehurek_lrec} with the same hyperparameters proposed in \cite{bruss2019deeptrax} and \cite{perozzi2014deepwalk}.


\subsubsection{External and Internal Evaluation}
\label{sub:extr}
In this work, we focus on evaluating embeddings both internally, meaning their ability to capture graph structure and externally, meaning their distinguishing power against node labels.
In the embedding space, similar nodes are expected to be placed closer together, but the notion of similarity can be arbitrarily defined based on node features, neighborhoods or connectivity patterns. Communities are a broadly used way of graph partitioning and can capture complex structural similarity. Consequently, they make a good test case for evaluating how graph structural properties are represented in the embedding space. Two learning problems are generated from this: pairwise community detection, which is a binary classification task of whether a pair of nodes belong in the same community and node level community prediction, which we treat as a multi-class classification problem of predicting the community a node belongs in given its embedding representation. For graphs that contain node labels, like BlogCatalog and Flickr, we treat them the same way. The goal in both tasks is to test the embeddings' efficiency to separate nodes. For community detection we use Louvain's algorithm for optimizing modularity \cite{blondel2008fast}.

\begin{table}[]
\caption{Dataset Statistics}
\centering
\begin{tabular}{ccccc}
\hline
& \begin{tabular}[c]{@{}c@{}}Number of \\ nodes\end{tabular} & \begin{tabular}[c]{@{}c@{}}Number of \\ edges\end{tabular} & Density                     & \begin{tabular}[c]{@{}c@{}}Number of \\ communities\end{tabular}          \\ \hline
Brand Level Merchants & over 100,000                 & over $8 \times 10^{6}$ & $1.2 \times 10^{-3}$& \begin{tabular}[c]{@{}c@{}}400\\  (80 for 95\% of nodes)\end{tabular}   \\ \hline
BlogCatalog  & 10,312  & 333,983                & $6.3 \times 10^{-3}$ & 6 \\ \hline
Flickr  & 80,513 & 5,899,882 & $1.18 \times 10^{-3}$ & 17 \\ \hline
\end{tabular}
\label{tab:Stats}
\end{table}

\section{Results}
\label{Results}
\begin{figure}[h!]
  \centering
  \begin{subfigure}[b]{\linewidth}
    \includegraphics[width=\linewidth]{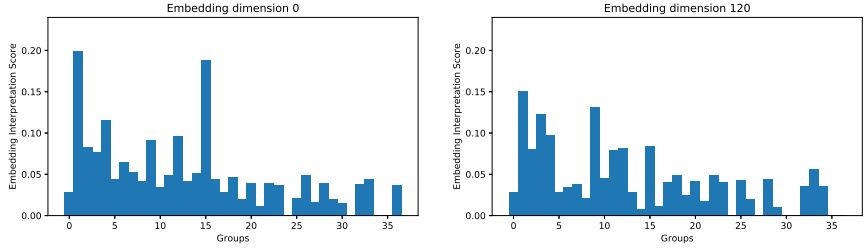}
    \caption{{\it IS} for the $0^{th}$ and $120^{th}$ embedding dimensions across node groups for BlogCatalog data (D = 128)}
  \end{subfigure}
  
  \begin{subfigure}[b]{\linewidth}
    \includegraphics[width=\linewidth]{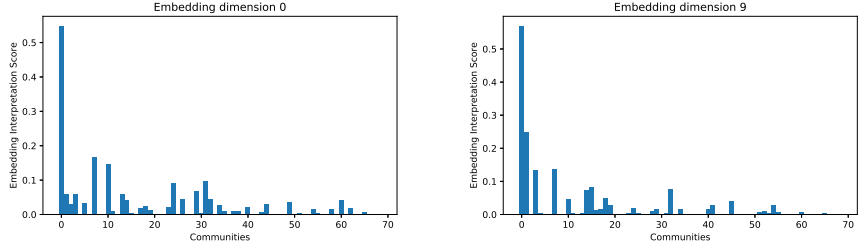}
    \caption{{\it IS} for the $0^{th}$ and $9^{th}$ embedding dimensions across communities for Brand Level Merchants (D = 10)}
  \end{subfigure}
  
  \begin{subfigure}[b]{\linewidth}
    \includegraphics[width=\linewidth]{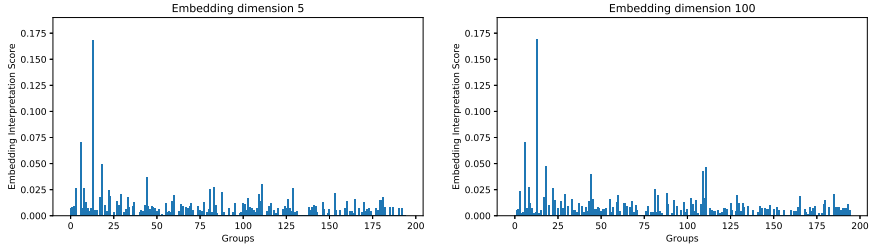}
    \caption{{\it IS} for the $5^{th}$ and $100^{th}$ embedding dimensions across communities for Flickr (D = 128)}
  \end{subfigure}
  \caption{Interpretability scores over node categorizations for selected embedding dimensions}
  \label{fig:IS}
\end{figure}

For our experiments we produced embeddings for all datasets with embedding dimensionality of 10, 64 and 128. First, we examine Interpretability Scores ({\it IS}) aggregated over different axes to explore the association of the embedding space with both external and internal node categorization. Figure \ref{fig:IS} shows the distribution of {\it IS} values over node communities for Brand Level Merchants and over node groups for the BlogCatalog and Flickr graphs. We observe that some node categories are highly associated with multiple dimensions of the embedding space (e.g. community 0 in Brand Level Merchants). These are the most highly populated categories and contain a larger variety of patterns expressed in multiple dimensions. Each embedding dimension appears to also be individually correlated with a particular subset of node categories. For instance the $0^{th}$ dimension for BlogCatalog is mostly correlated with groups 1 and 15, while the $120^{th}$ dimension carries information for groups 1, 14 and 3. 

\begin{table}[h!]
\centering
\caption{Performance for different classification tasks with various embedding dimensionality values. In binary classification values are {\it F1-scores}, in multi-class {\it micro-averaged F1} and LPAUC is Link Prediction {\it AUC}.} 
\begin{tabular}{ccccccccc}
\hline
\multirow{3}{*}{D} & \multicolumn{3}{c}{Brand Level Merchants}         & \multicolumn{5}{c}{BlogCatalog}    \\ \cline{2-9} 
& \multicolumn{3}{c}{Community} 
& \multicolumn{2}{c}{Group} 
& \multicolumn{3}{c}{Community}                         \\ \cline{2-9} 
& Binary    & Multi-class   
& \multicolumn{1}{c}{\begin{tabular}[c]{@{}c@{}}LP\\ AUC\end{tabular}} & Binary & Multi-label  & Binary     
& Multi-class  
& \multicolumn{1}{c}{\begin{tabular}[c]{@{}c@{}}LP\\ AUC\end{tabular}} \\ \hline
10 & \textbf{0.78} & 0.84 & \textbf{0.98} & 0.55 & 0.35 & 0.71 & \textbf{0.86} & 0.87 \\ \hline
64 & 0.71 & \textbf{0.86} & 0.95 & 0.75 & \textbf{0.42} & 0.68 & 0.80 & 0.90 \\ \hline
128 & 0.71 & 0.85 & 0.94 & \textbf{0.78} & 0.40 & \textbf{0.72} & 0.83 & \textbf{0.93} \\ \hline
\multirow{3}{*}{D} 
& \multicolumn{3}{c}{\multirow{2}{*}{}} 
& \multicolumn{4}{c}{Flickr}  &                         \\ \cline{2-9} 
\multicolumn{4}{c}{}
& Binary    & Multi-label  & Binary & Multi-class       
& \multicolumn{1}{c}{\begin{tabular}[c]{@{}c@{}}LP\\ AUC\end{tabular}} \\ \hline
\multicolumn{1}{c}{10}  & \multicolumn{1}{c}{--} 
& \multicolumn{1}{c}{--} & --                            & \multicolumn{1}{c}{\textbf{0.70}} & \multicolumn{1}{c}{0.37} 
& \multicolumn{1}{c}{\textbf{0.80}} & \multicolumn{1}{c}{0.85} 
& 0.95 \\ \hline
\multicolumn{1}{c}{64}  & \multicolumn{1}{c}{--} 
& \multicolumn{1}{c}{--} & --    
& \multicolumn{1}{c}{\textbf{0.70}} & \multicolumn{1}{c}{\textbf{0.40}} & \multicolumn{1}{c}{0.70} & \multicolumn{1}{c}{0.88} & \textbf{0.96} 
\\ \hline
\multicolumn{1}{c}{128} & \multicolumn{1}{c}{--} & \multicolumn{1}{c}{--} & --      
& \multicolumn{1}{c}{0.67} & \multicolumn{1}{c}{\textbf{0.40}} & \multicolumn{1}{c}{0.77} & \multicolumn{1}{c}{\textbf{0.94}} 
& \textbf{0.96} \\ \hline
\end{tabular}
\label{tab: F1}
\end{table}
Next, we report in Table \ref{tab: F1} the performance of different dimensionality embeddings on a set of prediction tasks described in Section \ref{sub:extr}. We observe that, by increasing the number of embedding dimensions, the ability to predict community membership does not improve, with an edge given to denser representations in BlogCatalog and Brand Level Merchants. Interestingly, performance in all node classification tasks we undertook is highly linked with Interpretability Scores distribution (see Figure \ref{fig:IS}), with the highest values being achieved for community prediction over node classification.
Performance in external node classification increases with the number of dimensions for the BlogCatalog data, while for Flickr data medium sized embeddings outperform the rest in this task. We can conclude that hyperparameter tuning can be based on two axes: graph properties of the dataset and the structures of the original graph we need the embeddings to capture.

Link prediction accuracy, for which random-walk based approaches optimize, appears to be correlated with external node classification. This is not always the case with community prediction, which favors smaller sized embeddings in BlogCatalog and Brand Level Merchants while link prediction improves with higher number of dimensions in the same datasets. In the Flickr graph different dimensionality embeddings achieve almost identical link prediction AUC scores, but show big deviations in performance in community prediction. Our findings imply that optimizing for link occurrence or external labels alone is not always sufficient to evaluate the embedding space as a whole and graph structure based tasks can shed light into the quality of latent representations. This is only the first the step in an effort to design a generalizable evaluation framework for different graph representation approaches across graphs with varying properties.

\bibliography{neurips}

\bibliographystyle{plain}
\end{document}